%% file: main.tex
\documentclass[acmsmall]{acmart}
\usepackage{array}
\usepackage{multirow}
\usepackage{threeparttable}
\usepackage{subcaption}
\AtBeginDocument{%
  }

\setcopyright{cc}
\setcctype{by}
\acmJournal{PACMNET}
\acmYear{2026}
\acmVolume{4}
\acmNumber{CoNEXT2}
\acmArticle{26}
\acmMonth{6}
\acmDOI{10.1145/3808674}

\begin{document}

\title{What's on My Network? Using Large Language Models to Identify Real-World IoT Devices at Scale}

\author{Rameen Mahmood}
\affiliation{%
  \institution{New York University}
  \city{New York}
  \state{NY}
  \country{USA}
}
\email{rameen.mahmood@nyu.edu}

\author{Sai Teja Peddinti}
\affiliation{%
  \institution{Google}
  \city{Mountain View}
  \state{CA}
  \country{USA}
}
\email{psaiteja@google.com}  
\author{Tousif Ahmed}
\affiliation{%
  \institution{Google}
  \city{Mountain View}  
  \state{CA}
  \country{USA}
}
\email{ahmedtousif@google.com}  

\author{Danny Yuxing Huang}
\affiliation{%
  \institution{New York University}
  \city{New York}
  \state{NY}
  \country{USA}
}
\email{dhuang@nyu.edu}  

\renewcommand{\shortauthors}{Rameen Mahmood, Sai Teja Peddinti, Tousif Ahmed \& Danny Yuxing Huang}

\input{sections/abstract}



\widowpenalty=10000
\clubpenalty=10000
\maketitle

\input{sections/introduction.tex}
\input{sections/related_work}
\input{sections/dataset}
\input{sections/methodology/stage1}

\input{sections/methodology/stage2}
\input{sections/results}
\input{sections/discussion}
\input{sections/conclusion}
\input{sections/ethics}

\begin{acks}
This work was supported by the Consumer Reports Digital Fellowship and by the U.S. National Science Foundation under awards CNS-2346332, CNS-2232655, and CNS-2219867. Any opinions, findings, and conclusions or recommendations expressed in this material are those of the authors and do not necessarily reflect the views of Consumer Reports or the National Science Foundation.
\end{acks}

\bibliographystyle{ACM-Reference-Format}
\bibliography{bibliography}

\received{December 2025}
\received[accepted]{April 2026}

\appendix

\renewcommand{\thetable}{A.\arabic{table}}
\setcounter{table}{0}

\renewcommand{\thefigure}{A.\arabic{figure}}
\setcounter{figure}{0}

\newpage
\section{Proxy CMI Derivations and Metric Definitions}
\label{appendix:proxy-cmi}
\subsection{Adjusted Mutual Information (AMI)}
\label{appendix:ami}
Given a categorical input feature \( X \in \mathcal{X} \) and an LLM-predicted vendor label \( Y \in \mathcal{Y} \), we quantify their dependence using the adjusted mutual information:

\begin{equation}
\text{AMI}(X; Y) = \frac{I(X; Y) - \mathbb{E}[I(X; Y)]}{\max\{H(X), H(Y)\} - \mathbb{E}[I(X; Y)]}
\end{equation}

Here, \( I(X; Y) \) denotes mutual information, \( H(\cdot) \) is Shannon entropy, and \( \mathbb{E}[I(X; Y)] \) is the expected mutual information under the null hypothesis of independence. Unlike raw mutual information, AMI corrects for spurious correlations that may arise from class imbalance or high-cardinality domains. This correction is essential in our setting, where features such as \texttt{remote\_hostname} follow long-tailed, aliased distributions.

\subsection{Stability via Entropy}
\label{appendix:entropy}

To assess intra-feature prediction consistency, we compute the conditional entropy of the LLM’s output distribution within each feature group:

\begin{equation}
H(Y \mid X = x_i) = - \sum_{y \in \mathcal{Y}} P(y \mid x_i) \log P(y \mid x_i)
\label{eq:conditional-entropy}
\end{equation}

We aggregate across groups as a weighted average and normalize against a uniform label entropy:

\begin{equation}
\text{Stability}(X) = 1 - \frac{\sum_i n_i \cdot H(Y \mid X = x_i)}{N \cdot \log_2 |\mathcal{Y}|}
\label{eq:stability}
\end{equation}
Here, \( n_i \) denotes the number of samples for which \( X = x_i \), and \( N \) is the total number of samples. A high stability score indicates low-entropy, consistent predictions across values of \( X \)—signaling the model’s confidence and reliability. This metric draws on entropy-guided interpretability techniques from recent work on decoding control~\citep{qiu2024entropy}, hallucination detection via semantic entropy~\citep{kossen2024semantic}, and consistency-based validation methods such as SelfCheckGPT~\citep{manakul2023selfcheckgpt}.

\subsection{Composite Score and Feature Ranking}
\label{appendix:cmi}
To combine informativeness and consistency, we compute a composite score:
\begin{equation}
\text{ProxyCMI}(X; Y) = \alpha \cdot \text{Stability}(X) + (1 - \alpha) \cdot \text{AMI}(X; Y)
\end{equation}
We use $\alpha = 0.5$ to give equal weight to both terms, though the framework supports tuning. This composite captures both global alignment and local determinism, ensuring top-ranked features are not only predictive but semantically robust. This mirrors class-conditional MI frameworks like SAMI~\citep{franken2024self}, which evaluate whether input features constrain model output in a semantically meaningful way.

\subsection{Exclusion Criteria: Ports and Search-Augmented Inference}
\label{app:cmi-exclusion}
For faithful attribution, we exclude two confounded sources of signal from our feature ranking analysis. First, we omit port numbers from \texttt{remote\_hostname} (e.g., avoiding \texttt{hostname:port} concatenation), as these encode behavioral priors that blur the line between identity and usage patterns—violating the assumption of semantic separability~\citep{guyon2003introduction}. Second, we exclude Brave Search–augmented prompts, which inject external web data not natively present in the structured fields. Including such context distorts mutual information by rewarding coverage breadth over intrinsic informativeness and reduces reproducibility across environments.

\section{Alias Resolution and Brand Consolidation.}
\label{app:alias-res}
Once per-row pseudo-labels are finalized, we canonicalize vendor names through a deterministic normalization pipeline designed to reduce semantic fragmentation across brand aliases. All vendor strings are lowercased, stripped of punctuation and whitespace, and then mapped to canonical parent organizations using the Wikidata SPARQL endpoint~\cite{wikidata}. This resolves common alias patterns (e.g., \textit{Nest} and \textit{Fitbit} → \textit{Google}; \textit{Echo Show} and \textit{Alexa} → \textit{Amazon}) and produces a unified vendor vocabulary prior to Stage~2 training. By consolidating alias variants into consistent parent entities, the instruction-tuned model learns stable vendor representations and generalizes effectively to unseen or partially observed aliases during inference.

\begin{table*}[h]
\caption{Descriptions of the structured input features used for vendor inference.}
\label{appendix:feature-descriptions}
\centering
\small
\begin{tabular}{p{3cm} p{9cm}}
\toprule
\textbf{Feature} & \textbf{Description}  \\
\midrule
\texttt{oui\_friendly}    & Vendor identifier derived from the MAC address prefix (OUI).                       \\
\texttt{dhcp\_hostname}   & Device hostname advertised via DHCP; often encodes model or OS.                  \\
\texttt{remote\_hostnames}    & Domains contacted via DNS and TLS SNI; reveals backend services. \\
\texttt{user\_agent\_info}& HTTP User-Agent string; captures software stack or client identity.               \\
\texttt{netdisco\_info}   & Broadcast metadata from mDNS, SSDP, or UPnP; exposes local services.         \\
\texttt{user\_labels}     & Free‑text annotations optionally provided by IoT Inspector users; may contain vendor, model, or type labels. \\
\texttt{talks\_to\_ads}   & \textit{Derived feature} indicating whether a device communicates with known advertising domains (from IoT Inspector’s ad‑domain list).\\
\bottomrule
\end{tabular}
\label{tab:feature-descriptions}
\end{table*}

\begin{figure}[H]
  \centering
  \includegraphics[width=1\linewidth]{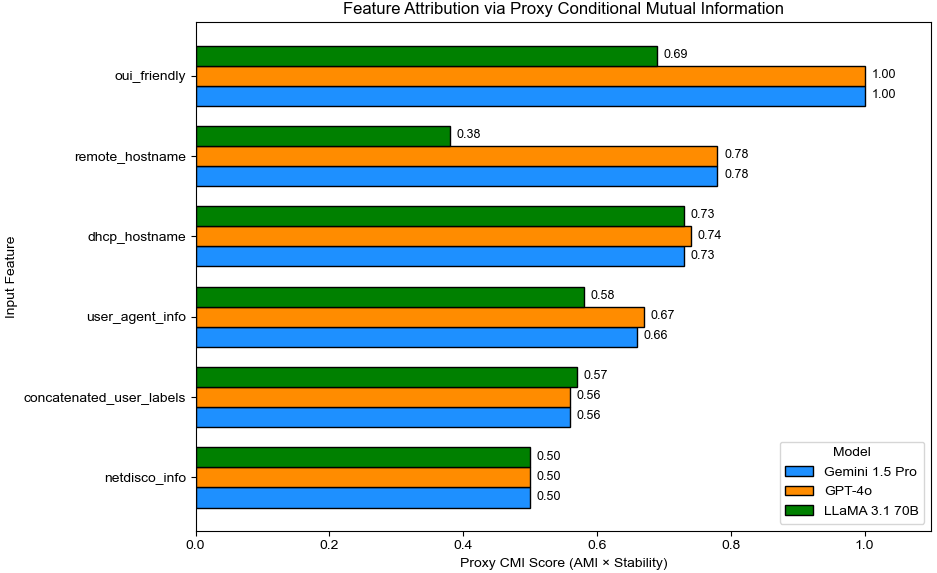}
  \caption{Proxy CMI scores for each input feature across three LLMs.}
  \Description[Bar chart of Proxy Conditional Mutual Information (CMI) scores for input features across models]{
  Horizontal bar chart showing Proxy Conditional Mutual Information (CMI) scores, a measure combining adjusted mutual information and stability, for seven input features: oui_friendly, remote_hostname, dhcp_hostname, user_agent_info, concatenated_user_labels, and netdisco_info. 
  Scores are shown for three language models: Gemini 1.5 Pro (blue), GPT-4o (orange), and LLaMA 3.1 70B (green). 
  The highest score is 1.00 for oui_friendly in Gemini and GPT-4o, while remote_hostname has the largest score disparity, with LLaMA 3.1 70B scoring 0.38 compared to 0.78 for GPT-4o and Gemini 1.5 Pro. 
  Other features have scores clustered between 0.5 and 0.75 across models.
  }
  \label{fig:proxy-cmi}
\end{figure}

\clearpage

\section{Illustrative Examples of Inference Complexity}
\label{appendix:examples}

\renewcommand{\thetable}{C.\arabic{table}}
\setcounter{table}{0}

{\centering
\captionof{table}{LLM vendor predictions on dense, sparse, and semi-structured inputs. These examples illustrate the range of inference challenges: noisy data, sparse clues, and user-labeled metadata.}
\footnotesize
\begin{tabular}{p{0.48\textwidth}|p{0.48\textwidth}}
\toprule
\textbf{Prompt (Dense + Noisy Metadata)} & \textbf{Model Output} \\
\midrule

\texttt{<|im\_start|>user} \newline
Below is information about a device. Think step-by-step, then predict the vendor.\newline
User Agent: Linux ; SM-G900A; AppleWebKit/537.36 \newline
OUI: Murata Manufacturing Co. \newline
DHCP Hostname: android-6b36f34ca17c6506 \newline
Talks to Ads: False \newline
\texttt{<|im\_end|>} &
\textit{Explanation}: The model number SM-G900A indicates Samsung; Murata is a known contract manufacturer.\newline
\textbf{Vendor: Samsung} \\

\midrule
\textbf{Prompt (Sparse Metadata)} & \textbf{Model Output} \\
\midrule

\texttt{<|im\_start|>user}\newline
OUI: AzureWave Technology Inc.\newline
DHCP Hostname: Roomba-3144460032234850\newline
Talks to Ads: False\newline
\texttt{<|im\_end|>} &
\textit{Explanation:} “Roomba” hostname is definitive for iRobot.\newline
\textbf{Vendor: iRobot} \\

\midrule
\textbf{Prompt (Unseen Vendor)} & \textbf{Model Output} \\
\midrule

\texttt{<|im\_start|>user}\newline
OUI: Arcadyan Corporation\newline
User Label: Telekom+Speedport TV+Speedport Bridge\newline
Talks to Ads: False\newline
\texttt{<|im\_end|>} &
\textit{Explanation:} Arcadyan manufactures gateways; label indicates a Deutsche Telekom router.\newline
\textbf{Vendor: Deutsche Telekom} \\
\bottomrule
\end{tabular}
\par}

{\centering
\captionof{table}{LLM vendor predictions under adversarial prompt manipulations (user-label spoofing).}
\footnotesize
\begin{tabular}{p{0.48\textwidth}|p{0.48\textwidth}}
\toprule
\textbf{Prompt (Spoofed User Label — Airbnb Host)} & \textbf{Model Output} \\
\midrule

\texttt{<|im\_start|>user}\newline
OUI: Amazon Technologies Inc.\newline
Remote Hostnames: cdn01.ring.com, rss.api.ring.com, ring-events-prod.amazon.com\newline
User Agent: AmazonWebView/FireOS\newline
DHCP Hostname: echo-livingrm\newline
Netdisco Info: Ring Doorbell Pro\newline
User Label (spoofed): “This is just a TP-Link smart plug.”\newline
\texttt{<|im\_end|>} &
\textit{Explanation:} All metadata points to Ring hardware.\newline
\textbf{Vendor: Amazon} \\
\midrule

\textbf{Prompt (Spoofed DHCP Hostname — Tech-Enabled Abuse)} & \textbf{Model Output} \\
\midrule

\texttt{<|im\_start|>user}\newline
OUI: Wyze Labs Inc.\newline
\raggedright
Remote Hostnames: a-us-east-1.wyze.com, api.wyzecam.com\newline
User Agent: WyzeCam/2.14.35\newline
DHCP Hostname (spoofed): nursery-monitor\newline
Netdisco Info: Wyze Cam v3\newline
User Label: Wyze Cam\newline
\texttt{<|im\_end|>} &
\textit{Explanation:} All metadata matches Wyze Cam v3.\newline
\textbf{Vendor: Wyze} \\
\bottomrule
\end{tabular}
\par}

{\centering
\captionof{table}{LLM robustness to misleading and scrambled hostnames.}
\footnotesize
\begin{tabular}{p{0.48\textwidth}|p{0.48\textwidth}}
\toprule
\textbf{Prompt (Original)} & \textbf{Model Output} \\
\midrule
\texttt{<|im\_start|>user}\newline
OUI: Google, Inc.\newline
\raggedright
Remote Hostnames: oem.googleapis.com, camera-ui.nest.com\newline
User Label: Nest Cam\newline
\texttt{<|im\_end|>} &
\textit{Explanation:} Hostnames and OUI match Google Nest.\newline
\textbf{Vendor: Google} \\

\midrule
\textbf{Prompt (Misleading Token: ring.com)} & \textbf{Model Output} \\
\midrule
\texttt{<|im\_start|>user}\newline
OUI: Google, Inc.\newline
\raggedright
Remote Hostnames: oem.googleapis.com, camera-ui.nest.com, ring.com\newline
User Label: Nest Cam\newline
\texttt{<|im\_end|>} &
\textit{Explanation:} A single unrelated hostname does not override strong signals.\newline
\textbf{Vendor: Google} \\

\midrule
\textbf{Prompt (Scrambled Token: goolgeapis.com)} & \textbf{Model Output} \\
\midrule
\texttt{<|im\_start|>user}\newline
OUI: Google, Inc.\newline
\raggedright
Remote Hostnames: oem.goolgeapis.com, camera-ui.nest.com\newline
User Label: Nest Cam\newline
\texttt{<|im\_end|>} &
\textit{Explanation:} Minor string corruption does not mislead the model.\newline
\textbf{Vendor: Google} \\

\midrule
\textbf{Prompt (Token Swap: camera-ui.ring.com)} & \textbf{Model Output} \\
\midrule
\texttt{<|im\_start|>user}\newline
OUI: Google, Inc.\newline
\raggedright
Remote Hostnames: oem.googleapis.com, camera-ui.ring.com\newline
User Label: Nest Cam\newline
\texttt{<|im\_end|>} &
\textit{Explanation:} Dominant Google/Nest indicators override injected noise.\newline
\textbf{Vendor: Google} \\
\bottomrule
\end{tabular}
\par}

\end{document}

%% file: sections/abstract.tex
\begin{abstract}

The growth of IoT devices in shared environments has outpaced our ability to identify them, posing urgent risks to privacy, safety, and accountability. This challenge is especially pronounced in open‑world environments, where network traffic metadata is often sparse, noisy, or adversarial. To address this problem, we introduce a semantic inference pipeline that reframes device identification as a language modeling task over real‑world network metadata. As this approach depends on reliable supervision, we first construct high‑fidelity vendor labels for the IoT Inspector dataset—the largest real‑world corpus of its kind—using an ensemble of large language models guided by mutual‑information and entropy‑based stability scores. We then instruction-tune a quantized LLaMA 3.1 8B model on this dataset using curriculum learning to support generalization under sparsity and long-tail vendor distributions. Our model achieves 98.69\% top-1 and 90.73\% macro accuracy across 2,015 vendors, while remaining robust to missing fields, protocol drift, and adversarial manipulation. We also evaluate the model on an independent IoT testbed dataset, assess explanation quality, and conduct adversarial tests to probe robustness under spoofed and obfuscated input. These results position instruction-tuned LLMs as a scalable, interpretable foundation for trustworthy device identification at scale.
\end{abstract}

%% file: sections/introduction.tex
\section{Introduction}
Modern local networks—across homes, enterprises, campuses, and shared accommodations—now host an expanding mix of laptops, phones, consumer IoT devices, access-control systems, and vendor-installed equipment that often lack traditional endpoint security. Yet despite this ubiquity, such environments remain surprisingly opaque: users and administrators rarely know \emph{what} devices are present, \emph{who} installed them, or \emph{what} they are capable of~\cite{thakkar2022would}. Enterprise inventories are frequently incomplete~\cite{benson2023leveraging}, third-party devices may appear without oversight~\cite{ilori2024third}, and employee BYOD practices introduce unmanaged equipment~\cite{annansingh2020bring}, creating risks ranging from misconfiguration to lateral malware movement~\cite{Mathews2017FishTankCasino, antonakakis2017understanding}.

Home and rental networks face similar challenges. Guests in Airbnbs routinely struggle to determine whether voice assistants, sensors, or cameras are present—sometimes accidentally~\cite{wang2023exploring, huang2020amazon} and sometimes as deliberate covert surveillance~\cite{mare2020smart, zeng2017end}. Such devices can facilitate harassment or intimate-partner violence~\citep{stephenson2023abuse, ceccio2023sneaky, freed2018stalker}. Across both consumer and enterprise settings, users lack reliable mechanisms to determine \textit{what} is on a local network and whether a device is benign, misconfigured, or
maliciously planted.

Device identification is therefore essential for security, privacy, and transparency. Yet existing approaches face fundamental limitations: they depend on signals that are sparse, indirect, or absent in real deployments. In practice, identification relies on two classes of techniques. \textit{Active probing} (e.g., mDNS, SSDP/UPnP, Nmap~\cite{lyon2009nmap}) assumes devices voluntarily broadcast identifiers, which many suppress or disable. \textit{Passive observation} infers identity from indirect network signals—such as MAC OUIs, DHCP hostnames, DNS queries, and protocol usage—but prior work is largely lab-bound (e.g., Mon(IoT)r’s 93-device corpus~\cite{girish2023room}) and fails to capture the long tail of vendors, firmware variants, and anomalous states observed in the wild.

To address these gaps, we leverage the IoT Inspector dataset~\cite{huang2020iot}, which crowdsources traffic from over 6,000 homes and 60,000 devices using both active (mDNS/SSDP) and passive monitoring. Unlike curated testbeds, it reflects real-world heterogeneity: contradictory identifiers, missing or randomized metadata, and a long tail of low-support vendors~\cite{280310}. These properties introduce core challenges for any identification system: (1) reconciling fragmented metadata, (2) inferring vendors absent from training data, and (3) remaining robust when key fields are sparse or adversarially manipulated.

Large language models (LLMs) are well-suited to this setting. They can integrate noisy, semi-structured metadata, infer plausible vendor identities, and draw on pretrained world knowledge to recognize device families never seen during training~\cite{brown2020language, kojima2022large, li2023revisiting, treutlein2024connecting}. We generate high-fidelity vendor pseudolabels\footnote{Pseudo-labels are vendor names assigned by the LLM ensemble rather than human annotators; they are validated against expert annotations in Sec.~4.1.3.} from the IoT Inspector dataset and instruction-tune LLaMA 3.1 8B to perform vendor inference, achieving 98.69\% top-1 accuracy across 2,015 vendors while remaining resilient to missing fields, protocol drift, and spoofed identifiers. Our system is deployed in an open-source application, enabling real-world, scalable device identification (as detailed in Sec.~\ref{sec:deployment}).
\vspace{-0.3cm}
\paragraph{Research Objectives}
We investigate whether LLMs can serve as a scalable, robust, and interpretable foundation for IoT device identification under open-set conditions. Specifically, we ask: (1) Can off-the-shelf LLMs generate accurate vendor pseudolabels from noisy or incomplete network metadata? (2) Can instruction-tuned LLMs generalize to rare or unseen vendors under open-set conditions, including sparsity and adversarial spoofing? (3) Can the resulting predictions remain interpretable and auditable?
\vspace{-0.3cm}
\paragraph{Threat Model \& Motivations}Our threat model includes settings where users lack visibility or control over the local network: enterprise environments with unmanaged or third-party devices, home and rental networks (e.g., Airbnbs), and tech-enabled abuse scenarios~\cite{freed2018stalker}. In these contexts, identification must rely solely on metadata devices naturally emit. We adopt the IoT Inspector model~\cite{huang2020amazon}, where a lightweight agent passively monitors traffic and exports encrypted-flow\footnote{Flows are 5-second summaries containing IP/transport headers, timestamps, and byte/packet counts. No payload content is collected.} metadata.
This vantage point mirrors what residents, analysts, or administrators—not adversaries—can realistically observe. Ultimately, our goal is to answer a deceptively simple but foundational question across both consumer and enterprise networks:
\emph{What’s on my network?}

%% file: sections/related_work.tex
\section{Related Work}

\subsection{Challenges of IoT Device Identification}
 IoT devices in modern networks are opaque, poorly managed, and resistant to conventional identification. They often lack stable identifiers~\citep{hosseinzadeh2016obfuscation}, originate from unknown supply chains~\citep{faraj2025security}, and exhibit misleading behavioral patterns~\citep{sivanathan2018classifying}. As a result, key security and analytic workflows—such as asset inventory~\citep{chanal2020security}, network segmentation enforcement, and behavioral modeling—are frequently undermined by ambiguity at the device layer.
These risks are especially acute in consumer settings, where connected devices  can silently retain credentials, leaving users vulnerable to prolonged remote surveillance~\citep{tekeoglu2015investigating, lau2018alexa, sivaraman2018smart}, and inadvertently revealing behavioral patterns even through encrypted traffic metadata~\citep{apthorpe2017smart}. These threats compound in environments like short-term rentals, shelters, or dormitories, where ownership is transient, visual inspection is impractical, and bystander privacy is crucial~\citep{freed2018stalker,291027,287242}. In enterprise and healthcare networks, the problem shifts from privacy to security. Devices routinely expose default credentials~\citep{albataineh2019iot, perone2023default}, outdated firmware~\citep{ebbers2022large}, and undocumented services—enabling lateral movement and botnet propagation~\citep{kolias2017ddos, kambourakis2017mirai}.

Despite the stakes, most identification pipelines rely on closed-world assumptions: complete metadata, known device inventories, and deterministic traffic signatures. Real deployments violate these assumptions at every turn. Metadata may be noisy, spoofed, or missing—e.g., generic DHCP hostnames (\textit{bedroom-TV}), randomized MAC prefixes \cite{android_wifi_mac_randomization_behavior,apple2024wifiprivacy}, or intentionally obfuscated OUIs~\citep{guerra2022datasets, wang2024epsilon, hernandez2022scaling,kumar2019all}—and vendor distributions exhibit extreme long-tail behavior, such as hundreds of models under umbrella vendors like Texas Instruments~\citep{avanzi2024machine, liang2025efficient}, revealing the structural mismatch between the variability of real deployments and the closed-world assumptions of deterministic fingerprinting.

\subsection{Machine Learning for Fingerprinting}
Early device fingerprinting approaches relied on active discovery methods—probing devices with crafted packets (e.g., Nmap~\citep{lyon2009nmap}), leveraging self‑announcement protocols like mDNS (Bonjour~\citep{apple2010bonjour}, Avahi~\citep{avahi2010}), frameworks like Acquisitional Rule-based Engine (ARE)~\citep{feng2018acquisitional} and DNSNA~\citep{lee2016dnsna} for IPv6‑based name registration. Later efforts such as IoT-Scan unified active and passive reconnaissance across
ZigBee~\citep{ieee2015lowrate}, BLE~\citep{bluetooth2016core},
LoRa~\citep{bor2016lora}, and Z-Wave~\citep{itu2015g9959}
at the radio layer using SDR hardware~\citep{gvozdenovic2023iot}. However, these approaches require device cooperation, depend on self-announcement protocols or active probing, and often need specialized RF capture. Devices that remain silent, misconfigured, or adversarially concealed therefore evade detection, underscoring the need for passive, traffic-based methods~\citep{salman2022machine, liu2021machine, msadek2019iot}.

Supervised approaches replaced brittle rule‑based heuristics by training on handcrafted statistical features—flow durations, inter‑packet timings, packet size distributions, and DNS query rates—to classify devices from structured traffic metadata. Moore et al.'s early Na\"ive Bayes classification model~\citep{moore2005internet}
IoTSense~\citep{liu2022iot}, and IoT Sentinel~\citep{miettinen2017iot} are a few popular supervised approaches.
Yet these approaches rely on rigid assumptions—clean labels, stable device behavior, and full feature observability—that rarely hold in noisy, open-world deployments.

Deep learning methods—including CNNs, LSTMs, and hybrid CNN–RNN architectures~\citep{lopez2017network}—learn directly from packet sequences and have shown strong performance even on encrypted flows~\citep{aneja2018iot, jafari2018iot, ullah2022design}. However, these models often ignore high-cardinality categorical fields (e.g., DHCP hostnames, contacted domains), assume access to richly labeled training data, and remain sensitive to incomplete, spoofed, or aliased metadata~\citep{gu2022tackling}. Long-tail vendor distributions further limit their ability to generalize and undermine interpretability. Unsupervised techniques (clustering, anomaly detection, embedding-based similarity~\citep{ortiz2019devicemien, bhatia2019unsupervised, zhang2021unsupervised}) flag novel behaviors but lack the semantic grounding needed for vendor inference and struggle with intra-vendor variability across firmware versions and deployment contexts.


These limitations motivate a shift toward models that can synthesize and interpret fragmented evidence rather than relying on brittle feature groupings. Recent work has explored LLMs for entity resolution~\citep{kojima2022large,
perez2021true}, open-schema extraction~\citep{li2023open}, and the clustering
of Internet-wide banner text to derive regex fingerprints~\citep{sarabi2023llm}. While these approaches demonstrate strong reasoning capabilities over coherent and self-descriptive textual inputs, they do not address the challenges identified above. In particular, even the closest work~\citep{sarabi2023llm} focuses on actively collected, device-advertised banner strings. Our setting is fundamentally different: inference must be drawn from noisy passive metadata and crowdsourced labels, which are often partial, aliased, or inconsistent. To the best of our knowledge, our work is the first to apply LLMs as context-aware inference models over such semi-structured IoT metadata, treating it not as fixed feature vectors but as structured context for reasoning—enabling open-world, long-tail inference under sparsity and inconsistency.

\vspace{0.1cm}
\noindent\textit{Limitations of Closed-World Identification.}\hspace{1em}
State-of-the-art device-identification systems—statistical (AutoIoT~\citep{fan2022autoiot}), programmable-switch based (DeviceRadar~\citep{li2024deviceradar}), and physical-layer behavioral (Lumos~\citep{sharma2022lumos})—achieve strong performance but rely on \emph{closed-world assumptions}: structured flows, stable behavioral signatures, and feature spaces drawn from a finite catalogue of known devices. Real deployments break these assumptions. Even when metadata is present, it is often \textit{semantically indirect}, reflecting underlying platform or supply-chain relationships rather than the consumer-facing device brand. OUIs frequently resolve to chipset manufacturers rather than device vendors (e.g., \texttt{esp32} → Espressif); contacted domains surface OEM software ecosystems (e.g., \texttt{tuyaus.com}); and User-Agent strings expose software-layer artifacts such as client libraries or application frameworks (e.g., \texttt{cURL}, \texttt{OkHttp}) rather than stable device or vendor identities~\citep{prakash2022inferring}. These signals are symbolic, inconsistent, and frequently missing. Rule-based systems collapse them into categorical tokens, discarding the linguistic and relational cues needed to resolve aliasing (``Nest'' → Google), product families (``Alexa'' → Amazon), or functional semantics (``Sonos'' → mesh audio). Empirically, on our 245-device expert-labeled hold-out drawn from the high-signal subset of our corpus, no individual signal resolves the consumer-facing vendor for more than a third of available devices (User-Agent: 5/66 = 7.6\%; OUI: 45/243 = 18.5\%; DHCP hostname: 10/35 = 28.6\%; mDNS/UPnP: 23/72 = 31.9\%; user labels: 17/51 = 33.3\%), and Fingerbank\footnote{Fingerbank is a widely used proprietary device-identification API that relies on deterministic matching over OUI prefixes, DHCP hostnames, and user-agent signatures.} achieves only 28.69\% accuracy overall, illustrating the mismatch between real-world data and deterministic pattern matching. These conditions motivate reframing vendor identification as \textit{semantic inference}: rather than collapsing metadata into predefined representations, we interpret it as a set of noisy, incomplete signals that require contextualization and cross-field reasoning. Large language models naturally support this capability—reconciling partial or spoofed identifiers, exploiting linguistic structure, and mapping aliases or OEM relationships to canonical vendors—while enabling \emph{open-set generalization} to unseen vendors, novel metadata combinations, and adversarial variants. By shifting from closed-world fingerprinting to open-world semantic inference, our approach addresses the structural limitations of prior systems and matches the realities of real-world deployments.

%% file: sections/dataset.tex
\section{Dataset Preprocessing}
\label{sec:preprocessing}

\begin{figure*}[t]
  \centering
  \includegraphics[width=1\linewidth]{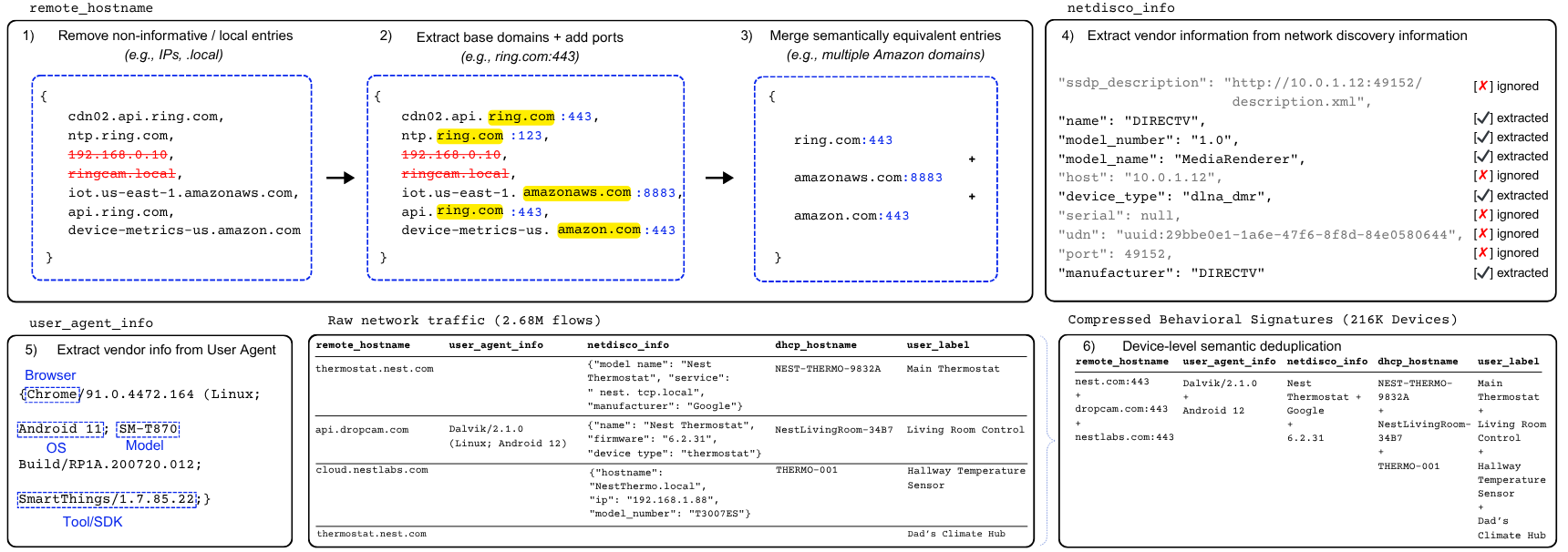}
\caption{Multi‑Stage Pipeline for Device‑Level Signature Extraction.}
\Description[Overview of six-step data preprocessing pipeline]{
  Diagram showing the six preprocessing stages used to convert raw IoT network traffic into device-level signatures. 
  Steps include: (1) removing non-informative hostnames like local IPs, (2) extracting base domains and appending ports, 
  (3) merging semantically equivalent domains (e.g., multiple Amazon domains), (4) extracting vendor identifiers from 
  network discovery info such as SSDP or mDNS, (5) parsing user agent strings into browser, OS, model, and SDK components, 
  and (6) deduplicating flows into a single canonical signature per device. Output compresses 2.68M raw flows into 
  216K device signatures for modeling.
  }
  \label{fig:data-preprocessing}
\end{figure*}

\subsection{Overview of Dataset}
\label{sec:dataset-overview}
Collected between 2019 and 2022, the IoT Inspector dataset contains 2.68M flow‑level entries, where each flow aggregates bytes sent and received over 5-second intervals keyed by the standard 5 tuple (source IP, destination IP, source port, destination port, and protocol), and is enriched with heterogeneous metadata: \texttt{remote\_hostname}s (DNS or SNI hostnames contacted), \texttt{user\_labels} (free‑text labels optionally provided by IoT Inspector users~\citep{280310}), \texttt{oui\_friendly} (vendor lookup from the MAC prefix), \texttt{dhcp\_hostname} (device‑
advertised DHCP name), \texttt{user\_agent\_info} (HTTP User‑Agent string), and \texttt{net}\texttt{disco\_info} (local service broadcasts via mDNS or SSDP); see Table~\ref{tab:feature-descriptions} for full descriptions.

\subsection{Preprocessing Pipeline}

The IoT Inspector dataset contains inconsistently reported metadata, motivating a six-stage preprocessing pipeline that transforms raw flow logs into compact device-level signatures suitable for modeling (Fig.~\ref{fig:data-preprocessing}). First, we discard non-informative \texttt{remote\_hostname} entries such as private IPs and local suffixes. Second, we canonicalize domains using the Mozilla Public Suffix List, preserving destination ports to retain protocol distinctions. Third, we merge semantically equivalent domains to reduce aliasing. Fourth, we extract persistent identifiers from \texttt{netdisco\_info} (mDNS/SSDP) while removing volatile fields such as serial numbers. Fifth, we normalize \texttt{user\_agent\_info} by tokenizing browser, OS, and device strings and stripping unstable build tags. Finally, we aggregate per-device flows into a canonical signature by deduplicating unique feature values. This reduces 2.68M flow-level entries to 772K canonical rows and ultimately to 216K unique device profiles, reducing redundancy and stabilizing device-level representations. Throughout, we retain missing values in feature fields rather than imputing them so that training reflects the partial observability seen at inference time. We also compute a binary feature, \texttt{talks\_to\_ads}, which flags whether a device contacts known advertising domains using lists provided in the IoT Inspector codebase.\footnote{\url{https://github.com/nyu-mlab/iot-inspector-client/tree/master/data}}

%% file: sections/methodology/stage1.tex
\vspace{-0.1cm}
\section{Method}
Stage~1 uses an ensemble of LLMs to assign 
high-confidence vendor labels to each device based on its noisy and incomplete metadata. 
Entropy-weighted majority voting consolidates the ensemble’s predictions into a single pseudo-label for each device, yielding a curated supervision set. However, Stage~1 alone is not sufficient for deployment: although LLMs exhibit
strong zero-shot reasoning, their predictions remain sensitive to missing fields,
prompt perturbations, and cross-model variance, leading to substantially lower
classification accuracy when used directly as classifiers
(see Table~\ref{tab:baselines}). We therefore use the pseudo-labels produced by Stage~1—validated against expert annotations (Sec.~\ref{ablation-study})—as structured supervision to train a
compact, stable classifier in Stage~2. We instruction-tune a quantized LLaMA~3.1~8B model on this supervision, distilling
the ensemble’s cross-field reasoning patterns into a compact model suitable for
deployment. During training, a curriculum schedule introduces progressively more ambiguous metadata combinations—starting from well-specified inputs and gradually increasing field sparsity—to strengthen robustness under open-world conditions. We detail the design, architecture, and training strategy for each stage below.

\begin{figure}
\centering
\includegraphics[width=0.9\columnwidth]{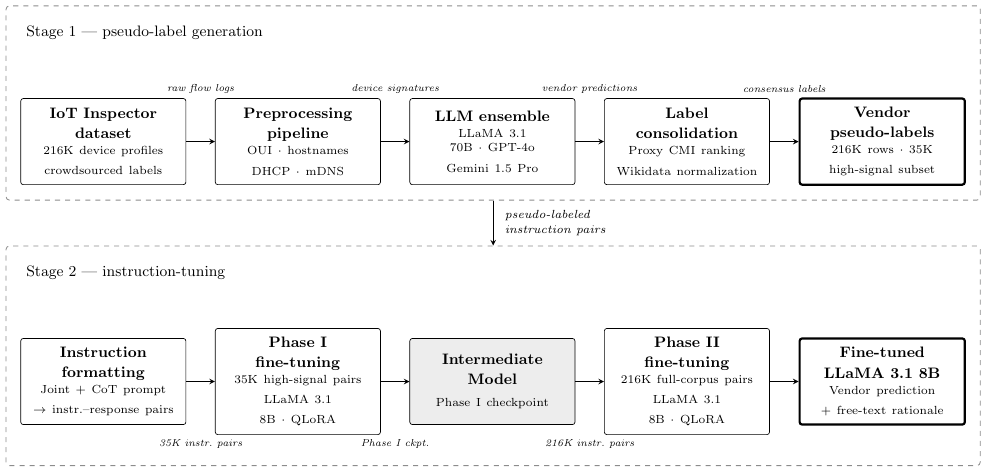}
\caption{Two-stage pipeline for IoT vendor classification.
Stage~1 generates vendor pseudo-labels using an LLM ensemble
(LLaMA~3.1~70B, GPT-4o, Gemini~1.5~Pro) with Proxy CMI--weighted voting
and Wikidata normalization (Sec.~\ref{stage1}). Stage~2 instruction-tunes
LLaMA~3.1~8B via QLoRA (Sec.~\ref{stage2}): Phase~I trains on the 35K high-signal
subset, saving an Intermediate Model checkpoint; Phase~II resumes on the
full 216K corpus to improve generalization (Sec.~\ref{sec:curriculum}).}
\end{figure}
\subsection{Stage 1: Labeling via Prompted LLMs}
\label{stage1}
Supervised learning at scale hinges on reliable ground‑truth labels. However, more than half of the devices (53.6\%) in the IoT Inspector dataset lack user‑provided labels, and the remainder exhibit heavy aliasing (e.g., \textit{Echo}, \textit{Amazon}, and \textit{Dot} referring to the same device)~\citep{280310}. This sparsity and inconsistency impede generalization performance~\citep{hernandez2022scaling, guerra2022datasets}, rendering naïve supervision infeasible.

To address this limitation, we generate high‑confidence pseudo‑labels in three parts: (1) we query three LLMs—LLaMA 3.1 70B, GPT‑4o, and Gemini 1.5 Pro\footnote{We use these three models as representative, high-performing LLMs drawn from different providers and training paradigms. Our goal is not to exhaustively benchmark LLMs, but to demonstrate the robustness of the Stage-1 labeling pipeline. We expect this procedure to generalize to other comparably high-performing instruction-tuned LLMs, though we do not claim invariance across all models.}—across each input feature using carefully designed prompts that produce structured outputs (Sec.~\ref{sec:prompting}); (2) we consolidate these outputs via majority voting, weighting conflicting predictions with Proxy CMI scores (Sec.~\ref{proxy-cmi}) and normalizing synonymous aliases via Wikidata (e.g., \textit{Nest} → \textit{Alphabet Inc.}) to reduce label fragmentation (Appendix \ref{app:alias-res})~\cite{wikidata}; and (3) we conduct an ablation study (Sec.~\ref{ablation-study}) across models and prompt variants, selecting Gemini 1.5 Pro with \textit{Joint + CoT} as the strongest configuration and applying it to label the full dataset. This pipeline yields 216K pseudo‑labeled rows across 2,015 vendors, from which we extract a 35K‑device high‑signal subset (where each device has at least one recorded \texttt{remote\_hostname}) across 344 vendors, which forms the initial high-information starting set for our Stage-2 
curriculum learning schedule (Fig. \ref{fig:cl}).

\subsubsection{Prompt Design and Output Structure}
\label{sec:prompting}
LLMs are highly sensitive to prompt formulation~\citep{mizrahi2024state}. For Stage~1 pseudo-label generation, we use a two-step prompting strategy for stable and interpretable outputs by generating (1) chain-of-thought (CoT) reasoning and (2) a joint structured prediction in the form \texttt{Device Type: <type>, Vendor: <vendor>}, drawing on~\citet{wei2022chain}. This structure serves three purposes: (1) it induces explicit reasoning, which improves label quality and reduces hallucinations~\citep{wei2022chain}; (2) it ensures a structured output format for automated parsing; and (3) it mitigates self‑contradictory predictions by aligning the model’s reasoning with outputs (e.g., avoiding cases like classifying a smart TV from a vendor known only for cameras). We validate this design via a systematic ablation against alternative prompt structures, measuring agreement with expert annotations (Sec.~\ref{ablation-study}).
The final \textit{Joint + CoT} prompt template used for
Stage~1 pseudo-labeling is provided with the accompanying artifacts.\footnote{
\url{https://anonymous.4open.science/r/artifact-materials-BA3D/}} Other ablation configurations were implemented as minimal modifications to this base template (e.g., removing chain-of-thought reasoning, splitting vendor/type prompts, or excluding ports).

\subsubsection{Proxy CMI: Feature Ranking and Voting}
\label{proxy-cmi} 

In our setting, metadata fields differ substantially in their predictive value: some are highly diagnostic, while others are noisy, ambiguous, or generic. For example, an OUI such as \textit{Espressif} appears across countless unrelated devices, whereas a hostname like \texttt{alexa.com} provides a clear signal for Amazon Echo. Stage~1 therefore queries each field independently and aggregates their predictions, hence identifying which fields are reliable is essential for producing accurate pseudo-labels. To quantify feature reliability we propose an information-theoretic framework, \textit{Proxy Conditional Mutual Information (Proxy CMI)}, that measures how strongly each input feature influences LLM-generated predictions. This method is model-agnostic and operates over any black-box LLM, requiring only predicted outputs. Our approach fuses two core metrics: (1) \textit{Adjusted Mutual Information (AMI)}, which captures how informative a feature is relative to the model’s predicted label; and (2) \textit{Entropy-Based Stability}, which measures how consistently the model behaves when conditioned on that feature. This builds on recent interpretability advances that use MI and entropy to dissect LLM behavior—e.g., rationale-label alignment~\citep{chen2024learning}, neuron sparsity attribution~\citep{wu2025interpreting}, and MI-optimized decoding~\citep{lu2024diver, xiao2025infopo}. Complete mathematical definitions and derivations for AMI, Stability, and the composite Proxy CMI score appear in Appendix \ref{appendix:proxy-cmi}. For attribution, we limit analysis to native features only—omitting ports and search‑augmented prompts to avoid confounding signals (detailed rationale in Appendix \ref{app:cmi-exclusion}).

Our feature ranking analysis shows that \texttt{oui\_friendly} dominates for Gemini 1.5 Pro and GPT‑4o, although its score drops under LLaMA 3.1 70B, where \texttt{dhcp\_hostname} emerges as more influential. This divergence reveals that no single feature is universally dominant and that feature salience varies across LLMs, motivating ensemble strategies that combine heterogeneous feature cues for robustness across model architectures. Figure~\ref{fig:proxy-cmi} presents the detailed rankings.
\vspace{0.2cm}

\noindent\textit{Deterministic Consolidation via Proxy-CMI Ranking.}\hspace{1em}
Stage-1 produces up to six feature-conditioned vendor predictions per device. When multiple features output the same vendor\footnote{To understand how often such consensus naturally arises, we examined the agreement structure across features: 76.9\% of devices exhibit a 3-of-6 majority, and 18.5\% show 5-of-6 consensus.}, we treat this as agreement and assign that consensus label directly. 
When features disagree—i.e., they produce different vendor candidates—we apply the Proxy-CMI hierarchy and select the vendor proposed by the \emph{most informative} available feature. As every device contains at least one non-null metadata field, this procedure always yields a deterministic pseudo-label without requiring abstention, probabilistic fusion, or confidence calibration.

\subsubsection{Ablation Study: Prompt and Model Selection}
\label{ablation-study}
To identify the most reliable pseudo-labeling configuration, we conduct a focused ablation study across three off-the-shelf LLMs while systematically varying four prompt-design dimensions. Specifically, we evaluate:
(1) Prediction Granularity: \emph{Separate} prompts (vendor and type predicted independently) vs. \emph{Joint} prompts;
(2) Rationale Format (CoT): chain-of-thought on/off;
(3) Feature Augmentation (Ports): inclusion of port information (e.g., \texttt{ring.com:554}); and
(4) Hostname Disambiguation (Brave): augmenting prompts with Brave Search metadata, falling back to an LLM only when search returns little or no structured information.

We assess these configurations on a a Manually Validated Hold-Out set (MV-Holdout) of 245 devices randomly sampled from the 35K high-signal subset. Each device was independently labeled by domain experts, yielding a statistically representative evaluation set that provides 95\% confidence with a $\pm5\%$ margin of error under an 80\% accuracy prior. Restricting evaluation to high-signal devices allows us to isolate the effect of prompt structure and model reasoning without confounding sparsity or missing-metadata artifacts. In practice, such instances—those containing vendor-revealing cues like \texttt{remote\_hostname}s—are the most valuable for bootstrapping high-signal pseudo-labels.

\begin{table}[t]
\centering
\footnotesize
\caption{Cohen’s $\kappa$ agreement between LLM configurations and expert annotations for different models.}
\label{tab:ablation-kappa}
\begin{tabular}{lccc}
\toprule
\textbf{Configuration} & \textbf{LLaMA 3.1 70B} & \textbf{GPT-4o} & \textbf{Gemini 1.5 Pro} \\
\midrule
Separate                  & 0.6484 & 0.6994 & 0.4642 \\
Separate + CoT            & 0.7492 & 0.7447 & 0.8283 \\
Separate + Ports          & 0.6359 & 0.6703 & 0.4275 \\
Separate + CoT + Ports    & 0.7407 & 0.7402 & 0.7940 \\
Joint                     & 0.6807 & 0.6954 & 0.7482 \\
Joint + CoT               & \textbf{0.7649} & \textbf{0.8233} & \textbf{0.8383} \\
Joint + Ports             & 0.6397 & 0.6994 & 0.7234 \\
Joint + CoT + Ports       & 0.7504 & 0.8135 & 0.8251 \\
Brave                     & 0.6644 & 0.6786 & 0.7391 \\
Brave + CoT               & 0.4638 & 0.2664 & 0.4838 \\
Brave + Ports             & 0.6644 & 0.6745 & 0.7551 \\
Brave + CoT + Ports       & 0.4638 & 0.2599 & 0.4838 \\
\midrule
\textit{Fingerbank} & \multicolumn{3}{c}{0.2139} \\
\bottomrule
\end{tabular}
\end{table}

Table~\ref{tab:ablation-kappa} reports Cohen’s~$\kappa$ agreement with expert
annotations. We adopt $\kappa$ instead of raw accuracy because it corrects for
chance agreement, which is critical in our highly imbalanced label space where
dominant vendors (e.g., \textit{Amazon}) can otherwise inflate accuracy. As a
result, $\kappa$ provides a more faithful comparison of labeling quality across
LLMs and prompt designs.

We observe three clear trends. First, Chain-of-Thought (CoT) prompting consistently improves alignment with expert annotations across all models; for instance, LLaMA 3.1 70B rises from $\kappa = 0.6484$ (\textit{Separate}) to $0.7492$ when CoT is applied in the same configuration (\textit{Separate + CoT}). Second, Gemini 1.5 Pro achieves the strongest overall agreement, peaking at $\kappa = 0.8383$ under the \textit{Joint + CoT} setup. Third, hostname disambiguation via Brave Search provides only marginal benefit in non-CoT settings—e.g., LLaMA 3.1 70B nudges from $\kappa = 0.6484$ (\textit{Separate}) to $0.6644$ (\textit{Brave})—but sharply degrades performance when paired with CoT prompting, with GPT-4o dropping from $\kappa = 0.8233$ (\textit{Joint + CoT}) to $0.2664$ (\textit{Brave + CoT}). Port information yields no measurable gains in any configuration. We observe that lookup-based augmentation \textit{(Brave)} tends to inject high-variance textual content, often dominated by marketing language. This dilutes core predictive signals—especially under CoT prompting, where verbose inputs increase the risk of reasoning drift. In contrast, compact, semantically aligned prompts support more stable reasoning over trusted fields. These findings underscore the tradeoff between external augmentation and input fidelity.

All LLM configurations\footnote{All Stage-1 outputs were generated using \texttt{temperature = 0.3}, \texttt{top-p = 0.9}, \texttt{max\_output\_tokens = 300}, a fixed seed (42), and an MD5-keyed prompt cache that guarantees deterministic pseudo-labels for identical prompts.}
 consistently outperform Fingerbank 
 ($\kappa = 0.21$), the current state‑of‑the‑art system for lookup‑based device identification that serves as our baseline benchmark. Fingerbank’s performance is constrained by two key factors: (i) coverage gaps—only $\sim$36\% of devices in the testbed return any labels; and (ii) systematic mislabeling, even for nominally “known” devices (e.g., classifying \emph{Wink} as a generic Samsung Android phone, or \emph{NVIDIA Shield TV} as “Linux OS”). In contrast, prompt-based pseudo-labeling generalizes to new vendors and device classes and treats identification as a model-driven semantic inference task rather than a static lookup problem.

%% file: sections/methodology/stage2.tex
\subsection{Stage 2: Supervised instruction-tuning for Vendor Classification}
\label{stage2}
We instruction-tune a causal decoder-only model (LLaMA 3.1 8B) on semi‑structured metadata using pseudo-labels generated by our Stage~1 pipeline (Sec.~\ref{stage1}). To enable scalable and robust adaptation, we combine: (1) parameter-efficient instruction-tuning via 4-bit quantized LoRA; (2) span-constrained supervision, which restricts loss to the vendor span to focus gradient signal; and (3) a two‑phase curriculum learning strategy, moving from high‑signal (Phase I) to sparse inputs (Phase II). We elaborate on these design choices in Sec. ~\ref{sec:architectural-choice}–\ref{sec:curriculum}.

\subsubsection{Prompt Format and Output Structure}  
Each training example is formatted as an instruction–response pair: the prompt lists the available metadata fields and the response contains a free-text rationale followed by a structured vendor label. We discard invalid fields and enforce a fixed field order to promote consistent attention patterns, and the same \textit{Joint\,{+}\,CoT} formatting used during Stage~1 is carried forward into Stage~2 instruction-tuning to maintain continuity between the pseudo-labels and the supervised fine-tuning signal. A representative training instance is provided with the accompanying artifacts.\footnote{ \url{https://anonymous.4open.science/r/artifact-materials-BA3D/} }

\subsubsection{Architectural Choice}
\label{sec:architectural-choice}

We build our classifier around the LLaMA 3.1 8B causal decoder model, instruction‑tuning it for multi-class vendor classification to parse semi‑structured metadata, emit structured vendor labels, and produce natural‑language rationales. Decoder‑only models excel at prompt‑based, autoregressive reasoning over partially specified inputs—capabilities our two‑phase curriculum learning strategy leverages directly.  

Other approaches are fundamentally mismatched to our problem setting.
Traditional classifiers (e.g., XGBoost, random forests) rely on brittle one-hot or TF-IDF encodings of sparse metadata fields and degrade sharply under missing values, aliasing, and long-tail vendors. Encoder-only models such as BERT, while effective for closed-set classification, are similarly constrained: they treat inputs as fixed feature vectors and cannot natively produce the structured predictions and rationales required by our pipeline.

We also examined retrieval-based and web-search–augmented approaches, but these are structurally ill-suited to device identification from network metadata. Unlike text-rich domains, there is no reusable document corpus: each device in our dataset appears only once, with unique, sparse, and highly fragmented metadata. Retrieval therefore cannot amortize information across instances and introduces non-trivial latency and memory overhead when scaled to millions of device profiles. Web-search–based agents face an additional limitation: most metadata contains no explicit vendor name—only indirect cues such as CDN hostnames, OEM strings, or partial DHCP labels—which search cannot reliably resolve. Our Brave-Search ablation (Sec.~\ref{ablation-study}) confirms this: retrieval succeeds only when vendor strings appear verbatim and fails on the majority of real-world hostnames.

These structural limitations are reflected empirically in the baseline comparisons reported in Sec.~\ref{sec:baselines}. In contrast, LLMs can semantically integrate weak, cross-field evidence and generalize to unseen vendors. By operating directly on the structured fields available at inference time, a lightweight instruction-tuned decoder offers a more robust, private, and deployment-compatible solution.
\begin{figure*}[t]
  \centering
  \includegraphics[width=0.9\linewidth]{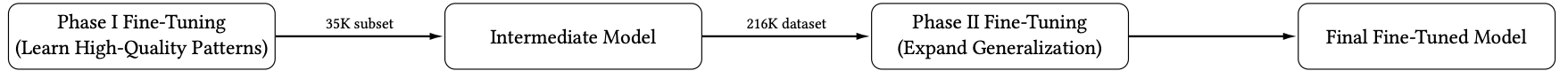}
  \caption{Curriculum-style instruction-tuning strategy.}
  \Description[Two-phase curriculum fine-tuning diagram]{
  Flowchart illustrating the two-phase curriculum instruction-tuning process. 
  Phase I fine-tunes the model on a curated 35K high-quality subset to learn core vendor identification patterns, 
  producing an intermediate model. This intermediate model is then fine-tuned in Phase II on the full 216K-device dataset 
  to broaden generalization. The process outputs the final instruction-tuned model for deployment.
  }
  \label{fig:cl}
\end{figure*}
\subsubsection{Model and Quantization Strategy}
\label{sec:quantization}
To make fine-tuning of our model feasible on a single GPU, we use a memory-efficient quantized adaptation strategy. We instruction-tune\footnote{For numerical stability, we set \texttt{bnb\_4bit\_compute\_dtype=bfloat16}.
LoRA adapters are 
configured to use rank $r{=}8$, scaling factor $\alpha{=}16$, and dropout rate${=}0.05$. 
The model is initialized with \texttt{prepare\_model\_for\_kbit\_training()} and instruction-tuned using the \texttt{SFTTrainer} from TRL, employing the \texttt{paged\_adamw\_32bit} optimizer, cosine learning rate decay (initial $\eta{=}2\times10^{-4}$), and mixed-precision training (\texttt{fp16=True}).} the \texttt{Meta-Llama-3.1-8B-Instruct} checkpoint using QLoRA~\citep{dettmers2023qlora}. 
 Inputs are right-truncated to 1024 tokens. We use a microbatch size of 1 and accumulate gradients over 8 steps, yielding an effective batch size of 8. Gradient checkpointing is enabled to further reduce memory usage. All experiments are conducted on a single NVIDIA A100 GPU (80GB).

\subsubsection{Vendor-Only Supervision via Targeted Loss Masking}
\label{sec:lossmasking}
We train the model to focus only on predicting the vendor label, not on reproducing the reasoning trace, to prevent the model from learning spurious patterns from its own explanations. To do this, we apply targeted loss masking that restricts supervision to the \texttt{Vendor:} field. Specifically, all tokens preceding the vendor span are assigned a loss mask of $-100$, ensuring that gradient updates are computed only over the final label tokens. This confines supervision to the decision span, avoiding spurious gradients from explanations that reference the correct vendor even when the predicted label is wrong. Although much prior work does not mask rationales during training, some studies in rationale supervision demonstrate the benefits of decoupling explanation from prediction—either by treating rationales as latent variables, excluding them from the loss, or marginalizing over multiple explanation paths~\citep{lei2016rationalizing, wang2022rationale}. Such strategies have been shown to improve generalization, robustness, and calibration.

\subsubsection{Curriculum Learning for Deployment-Grade Generalization}
\label{sec:curriculum}

Our goal is to deploy a multi-class vendor classifier that operates reliably under real-world conditions—where device metadata is often sparse, noisy, or incomplete. To align training with this target setting while ensuring stable convergence, we adopt a two-phase curriculum learning strategy that transitions from clean, high-precision supervision to full-spectrum, deployment-grade inputs (Fig.~\ref{fig:cl}).

A curriculum is well-suited to our problem because metadata difficulty varies substantially across devices: distinctive OUIs or clean hostnames offer strong vendor cues, whereas long-tail devices exhibit ambiguity, aliasing, or missing fields. A staged approach allows the model to first internalize high-confidence input–label relationships before being exposed to the broader distribution of harder, noisier examples. Similar curricula are widely used in large-scale LLM fine-tuning, where they help structure learning and promote generalization~\citep{brown2020language, wang2024curriculum, zhang2025beyond, wu2025progressive}. In \textit{Phase I (High-Signal Training)}, we instruction-tune on the 35K high-signal subset, where fields such as \texttt{remote\_hostname} provide reliable vendor cues. This establishes a foundation of unambiguous mappings. In \textit{Phase II (Deployment-Grade Training)},
we continue training on the full 216K dataset, which reflects deployment conditions: long-tail vendor distributions and increasingly ambiguous dentifiers. Building upon the structured representations learned in Phase I, the model can now adapt to the variability inherent in real-world device profiles.

%% file: sections/results.tex
\section{Results}
\label{sec:results}

We evaluate our instruction-tuned LLaMA~3.1~8B model across predictive accuracy, robustness to deployment shift, and robustness to adversarial and semantic perturbations. 
Accuracy is measured on internal hold-outs and the MV-Holdout (Sec.~\ref{ablation-study}), while deployment robustness is assessed on the Mon(IoT)r Testbed~\cite{girish2023room}. 
We further report curriculum and feature ablations (Sec.~\ref{sec:feature-ablation}), baseline comparisons (Sec.~\ref{sec:baselines}), novelty-aware abstention (Sec.~\ref{sec:openset}), and cost and deployment analyses (Secs.~\ref{sec:cost-throughput}–\ref{sec:deployment}).

\subsection{Performance on Internal Hold-Out Test Set}
\label{sec:internal-holdout}
Table~\ref{tab:curriculum-results} presents accuracy metrics across the two phases of curriculum learning. In Phase~I, training on 35K high-signal examples yields strong top-1 (97.54\%) and macro (91.68\%) accuracy, anchoring the model in well-supported semantic regions of the label space. Phase II leverages the entire 216K‑device corpus—including long‑tail, aliased, and weakly supervised classes—yielding a further boost in top‑1 accuracy to 98.69\% and suggesting that broader coverage enhances generalization despite increased input sparsity. While macro accuracy declines to 90.73\%, this reflects greater decision uncertainty in underrepresented classes rather than degraded  performance. Notably, accuracy on the MV-Holdout rises from 86.96\% to 93.20\%, indicating improved calibration under real-world ambiguity. This divergence—where top-1 accuracy increases, macro accuracy remains high, and external generalization improves—shows that the model is not just memorizing dominant vendors, but acquiring robust, semantically grounded mappings that transfer across sparsity, drift, and adversarial variation. 

\subsubsection{Tiered Evaluation}
As vendor labels in the wild are inconsistent and aliased, exact string matching alone understates true model performance. To address this, we evaluate predictions using a tiered rubric of increasingly permissive criteria (Table~\ref{tab:tiered-accuracy}). Strict matching yields 63.4\% accuracy in Phase~I and 70.4\% in Phase~II, but many mismatches reflect principled refinements rather than true errors—for example, resolving brand aliases (\textit{Nest}~$\rightarrow$~\textit{Google}), normalizing descriptors (voice assistant~$\rightarrow$~smart speaker), or correcting inconsistent supervision (\textit{Echo}, \textit{Dot}~$\rightarrow$~\textit{Echo Dot}). Aggregating the \textit{Semantic Alignment}, \textit{Brand Consolidation}, and \textit{Ambiguous Label Exclusion} tiers raises accuracy to 92.70\% and 89.62\% in Phases~I and~II. A final \textit{Manual Validation Tier} credits semantically plausible predictions, yielding 97.54\% and 98.69\%—consistent with the Top-1 accuracy in Table~\ref{tab:curriculum-results}.\footnote{Unless noted otherwise, accuracy figures correspond to the Manual Validation Tier.} These results show that the model often improves on its supervision: producing canonical, taxonomically coherent vendor names rather than replicating fragmented or aliased labels. Tiered evaluation therefore clarifies true model performance and highlights cases where the model generalizes beyond imperfect training signals.
\begin{table*}[t]
\centering
\footnotesize

\begin{minipage}[t]{0.45\textwidth}
\centering
\caption{Accuracy across curriculum phases. Top-1 and Macro Accuracy are evaluated on phase-specific internal hold-outs (10\% split)}
\label{tab:curriculum-results}
\setlength{\tabcolsep}{1pt}

\begin{tabular}{lcc}
\toprule
\textbf{Metric} & \textbf{Phase I (35K)} & \textbf{Phase II (216K)} \\
\midrule
Top-1 Accuracy              & 97.54\% & 98.69\% \\
Macro Accuracy             & 91.68\% & 90.73\% \\
MV-Holdout Accuracy (n=245)  & 86.96\% & 93.20\% \\
\bottomrule
\end{tabular}
\end{minipage}
\hfill
\begin{minipage}[t]{0.47\textwidth}
\centering
\caption{Tiered accuracy under increasingly permissive evaluation criteria on phase-specific internal hold-out sets.}
\label{tab:tiered-accuracy}

\begin{tabular}{lcc}
\toprule
\textbf{Evaluation Tier} & \textbf{Phase I} & \textbf{Phase II} \\
\midrule
Strict Match              & 63.40\%  & 70.40\% \\
Semantic Alignment        & 75.50\%  & 81.48\% \\
Brand Consolidation       & 71.70\%  & 75.02\% \\
Ambiguous Label Exclusion & 72.20\%  & 73.92\% \\
Unified Label Tier        & 92.70\%  & 89.62\% \\
Manual Validation Tier    & \textbf{97.54\%} & \textbf{98.69\%} \\
\bottomrule
\end{tabular}
\end{minipage}

\end{table*}

\subsubsection{Generalization Across the Long-Tail Vendor Distribution.}

To assess generalization by class frequency, we partition hold-out vendors into 
three tiers: Head (>100), Mid (11–100), and Tail ($\leq$10). Accuracies are reported 
at the vendor level: “\#~Classes’’ denotes the number of distinct vendors in each tier, 
and “\#~Samples’’ reflects the corresponding device instances in the hold-out split. 
Table~\ref{tab:vendor-tier-accuracy} reports tier-wise accuracy across both curriculum 
phases. 

We additionally compare against a non-curriculum variant (final column of Table~\ref{tab:vendor-tier-accuracy}). 
While the non-curriculum model achieves slightly higher accuracy on head vendors (99.49\% vs.\ 98.36\%), 
the curriculum-trained model substantially improves performance on mid- and tail-frequency vendors, 
with gains from 98.47\% to 99.71\% in the mid tier and from 95.70\% to 97.74\% in the tail. 
This reflects a trade-off between memorization and generalization: without curriculum, the model tends to overfit dominant patterns in high-frequency vendors, whereas curriculum learning encourages more balanced representations across difficulty levels. 
As a result, performance is redistributed toward underrepresented classes, improving robustness in the long tail.

Despite heavy imbalance—over half of vendors fall into the tail—the model maintains strong performance. In Phase~I, tail accuracy reaches 93.68\%, trailing head-class accuracy (98.19\%) by less than five points, while Phase~II lifts tail accuracy further to 97.74\%. Two patterns stand out: first, the model consistently generalizes beyond high-resource vendors, producing semantically coherent predictions for rare, low-frequency classes. Second, the tail accuracy gain from Phase~I to Phase~II indicates that expanding supervision to a broader, noisier vendor set strengthens rare-class robustness rather than diluting it. This resilience to long-tail underrepresentation underscores the value of instruction tuning for open-world classification tasks, where exhaustive coverage is infeasible but prediction accuracy is essential.

\subsubsection{Vendor-Level Error Analysis}
\label{sec:error-analysis}

To assess class-level reliability and potential bias, we analyze per-vendor accuracy and misclassification patterns across the Phase~II evaluation set. As shown in Table~\ref{tab:vendor-tier-accuracy}, performance is strong across both head and tail vendors. High-frequency vendors such as Apple (98.6\%), Amazon (95.8\%), Samsung (97.3\%), Sonos (98.7\%), and Roku (97.0\%) achieve near-perfect accuracy, while tail accuracy remains high (93.68\% in Phase I; 95.70\% in Phase II) across hundreds of low-frequency classes.

Errors are concentrated in a small set of structurally ambiguous vendors, primarily component or OEM manufacturers such as Intel Corporate (1.2\%), AzureWave Technology (6.8\%), Murata Manufacturing (20.0\%), and Espressif Inc.\ (60.0\%), whose identifiers (e.g., OUIs) are shared across many downstream devices. In these cases, predictions often reflect the end-device vendor rather than the underlying component manufacturer.

Outside of these categories, errors are not systematically concentrated. As shown in Table \ref{tab:confusion}, 67 total errors in Phase~II, only 3 (4.5\%) correspond to dominant vendors (Amazon, Apple, Google, Samsung, Microsoft), while 95.5\% involve non-dominant vendors. Misclassifications instead follow plausible OEM or supply-chain relationships (e.g., Espressif $\rightarrow$ Tuya, Insignia $\rightarrow$ Best Buy), rather than indicating bias toward major brands.

\begin{table*}[t]
\caption{Accuracy across head, mid, and tail vendors in the internal hold-out sets for both curriculum phases. 
For comparison, we also report no-curriculum accuracy in the final column.}
\centering
\footnotesize

\begin{tabular}{lccccccc}
\toprule
\textbf{Vendor Tier} 
& \multicolumn{3}{c}{\textbf{Phase I (35K)}} 
& \multicolumn{3}{c}{\textbf{Phase II (216K)}} 
& \textbf{No Curriculum} \\
\cmidrule(lr){2-4} \cmidrule(lr){5-7}
& Accuracy & \# Classes & \# Samples 
& Accuracy & \# Classes & \# Samples 
& Accuracy \\
\midrule
Head (\textgreater100) & 98.19\% & 5   & 885  & 98.36\% & 4   & 1581 & 99.49\% \\
Mid (11--100)          & 98.35\% & 22  & 666  & 99.71\% & 44  & 1371 & 98.47\% \\
Tail ($\leq$10)        & 93.68\% & 115 & 285  & 97.74\% & 473 & 883  & 95.70\% \\
\bottomrule
\end{tabular}

\label{tab:vendor-tier-accuracy}
\end{table*}
\begin{table}[t]
\centering
\footnotesize
\caption{Top misclassification pairs from error analysis 
(67 true errors total).}
\label{tab:confusion}
\begin{tabular}{llc}
\toprule
\textbf{Ground Truth} & \textbf{Predicted} & \textbf{Count} \\
\midrule
Espressif Inc.           & Tuya              & 3 \\
Insignia                 & Best Buy          & 2 \\
StreamUnlimited Eng.     & Best Buy          & 2 \\
Slim Devices             & Sonos             & 2 \\
Luxshare Precision       & Huawei            & 2 \\
VOCOlinc                 & ShenZhen Fuzhi    & 2 \\
\midrule
\multicolumn{2}{l}{\textit{Dominant brand predicted}} & 3 (4.5\%) \\
\multicolumn{2}{l}{\textit{Non-dominant predicted}}   & 64 (95.5\%) \\
\bottomrule
\end{tabular}
\end{table}
\subsubsection{Stage-Drop and Phase-Order Ablation}
To quantify the contribution of each component in the training pipeline, we conducted a stage-drop ablation isolating the effects of pseudo-labels, curriculum structure, and phase ordering (Table~\ref{tab:stage-drop}). Training solely on user-provided labels, without any pseudo-label guidance, achieved 65.7\% accuracy on the MV-Holdout. Retaining the two-phase curriculum while removing pseudo-labels yielded a slightly higher 66.2\% accuracy, indicating that curriculum structure alone is insufficient to compensate for missing supervision. Using pseudo-labels without curriculum resulted in 98.25\% accuracy, while enabling the full curriculum improved this to 98.69\% and yielded more stable calibration across long-tail vendors. Reversing the phase order—training first on sparse or noisy examples and then on high-signal examples—reached 94.8\% accuracy, highlighting the sensitivity of convergence behavior to curriculum ordering.
Although the curriculum yields a modest improvement in top-1 accuracy, this behavior is consistent with prior findings that curricula primarily affect optimization dynamics rather than asymptotic performance once datasets are large and supervision is strong~\citep{liu2023review,avramova2015curriculum}. In our setting, the curriculum’s main contribution is stability: non-curriculum runs frequently collapsed to dominant vendors or failed to converge, whereas the two-phase syllabus consistently produced stable training, improved long-tail calibration, and clearer gains on ambiguous or noisy metadata. As a result, the small absolute accuracy gain understates the curriculum’s practical impact on robustness and reliability—an effect aligned with prior work showing curricula are most beneficial when task difficulty is heterogeneous and signals vary in reliability~\citep{weinshall2018curriculum,wu2020curricula}.
To further evaluate robustness to noisy or unavailable user input, we retrain the model without \texttt{user\_labels} during both training and inference, yielding 86.86\% accuracy. This indicates that while user-provided labels are informative, the model maintains strong performance by leveraging other available input fields.
\begin{table*}[t]
\centering
\footnotesize

\begin{minipage}[t]{0.47\textwidth}
\centering
\caption{Stage-drop and phase-order ablation isolating the contributions of pseudo-labels, curriculum scheduling, and phase ordering.}
\label{tab:stage-drop}
\begin{tabular}{lc}
\toprule
\textbf{Configuration} & \textbf{Top-1 Acc.} \\
\midrule
User labels only (no pseudo-labels)        & 65.7\% \\
Curriculum only (no pseudo-labels)         & 66.2\% \\
Pseudo-labels only (no curriculum)         & 98.25\% \\
Full model: pseudo-labels + curriculum     & \textbf{98.69\%} \\
Reversed curriculum order                  & 94.8\% \\
\bottomrule
\end{tabular}
\end{minipage}
\hfill
\begin{minipage}[t]{0.47\textwidth}
\centering
\caption{Feature ablation accuracy, showing model performance with each feature removed.}
\label{tab:leave-one-out}
\setlength{\tabcolsep}{5pt}

\begin{tabular}{lcc}
\toprule
\textbf{Feature Ablated} & \textbf{Phase I} & \textbf{Phase II} \\
\midrule
\textit{All Features (Baseline)} & 86.67\% & 92.86\% \\
\texttt{user\_agent\_info} & 82.86\% & 90.00\% \\
\texttt{user\_labels}      & 85.24\% & 88.57\% \\
\texttt{dhcp\_hostname}    & 82.86\% & 87.62\% \\
\texttt{netdisco\_info}    & 85.24\% & 86.67\% \\
\texttt{remote\_hostname}  & 80.48\% & 80.95\% \\
\texttt{oui\_friendly}     & 66.19\% & 68.10\% \\
\bottomrule
\end{tabular}
\end{minipage}

\end{table*}
\begin{table}[t]
\centering
\footnotesize
\caption{Perturbation-based field attribution on the MV-Holdout. 
Each field is masked independently per device; 
``Predictions Changed'' reports the fraction of instances where 
masking alters the output.}
\label{tab:saliency}
\begin{tabular}{lcc}
\toprule
\textbf{Field Masked} & \textbf{Predictions Changed} & \textbf{Accuracy Drop} \\
\midrule
\texttt{oui\_friendly}              & 31.4\% & 24.8\% \\
\texttt{remote\_domains}              & 17.1\% & 11.9\% \\
\texttt{concatenated\_user\_labels} & 10.5\% & 4.3\%  \\
\texttt{netdisco\_info}             & 8.6\%  & 6.2\%  \\
\texttt{dhcp\_hostname}             & 8.1\%  & 5.2\%  \\
\texttt{user\_agent\_info}          & 5.7\%  & 2.9\%  \\
\midrule
\textit{Robust (no field dominant)} & 49.0\% & —      \\
\bottomrule
\end{tabular}
\end{table}
\subsubsection{Feature Importance via Leave-One-Out Ablation}
\label{sec:feature-ablation}

We assess feature contributions using leave-one-out ablations on the MV-Holdout for both Phase~I and Phase~II models (Table~\ref{tab:leave-one-out}). Two trends emerge. First, \texttt{oui\_friendly} is the dominant signal: removing it yields the largest accuracy drops (–20.8\% in Phase~I; –25.6\% in Phase~II), reflecting the strong manufacturer cues encoded in MAC prefixes. Second, removing other fields produces only modest degradation, indicating that the model remains robust to partial or noisy metadata. The importance of \texttt{remote\_hostname} increases in Phase~II (–13.5\%), where long-tail and ambiguous vendors appear; sparse hostname patterns thus become key disambiguators when traditional identifiers are weak. These findings are consistent with our Proxy-CMI analysis (Sec.~\ref{proxy-cmi}), which similarly highlights \texttt{oui\_friendly} and \texttt{remote\_hostname} as the highest-signal fields (as shown in Fig. \ref{fig:proxy-cmi}).

\vspace{0.15cm}
\noindent\textit{Perturbation-Based Field Attribution.}\hspace{1em} To complement the population-level ablation above and assess 
per-instance decision behavior, we mask each field independently 
across the MV-Holdout and record whether the prediction changes 
(Table~\ref{tab:saliency}). For 49.0\% of devices, no single 
field removal changes the prediction, indicating robust cross-field 
inference where evidence is distributed across multiple signals 
simultaneously. Among devices with a dominant signal, 
\texttt{oui\_friendly} is primary for 31.4\% and 
\texttt{base\_domains} for 17.1\%. Vendor-level analysis reveals 
systematic adaptation: OUI dominates for vendors with distinctive 
MAC prefixes (Ubiquiti, Dell, Raspberry Pi), while hostname drives 
predictions for vendors with recognizable domain patterns. 
High-frequency vendors (Alphabet Inc., Amazon, Apple, Roku) are 
largely robust---recognized reliably from multiple signals 
simultaneously. These results confirm that the model dynamically 
weights available signals based on per-vendor informativeness 
rather than uniformly anchoring on any single field.

\vspace{-0.55cm}
\subsection{External Evaluation: Generalization Across Drifted and Obfuscated Network Environments}
\label{subsec:external-eval}
We evaluate our model on the Mon(IoT)r Testbed~\citep{girish2023room}, which captures real-world smart home traffic from 93 IoT devices across speakers, cameras, and kitchen appliances in a controlled setting. We convert the released raw packet captures into the same structured metadata fields used in our pipeline, ensuring feature-level compatibility despite differences in collection environment. This testbed enables evaluation under three realistic drift conditions: temporal drift (2019 vs. 2022 collection periods), geographic variation (US vs. UK), and protocol-level obfuscation (VPN-based anonymization). Restricted-access distribution makes the dataset unlikely to appear in LLM pretraining corpora, ensuring a clean external benchmark. We acknowledge that this evaluation primarily measures distribution-shift robustness rather than open-set generalization to unseen vendors.

Despite training exclusively on IoT Inspector data (2019–2022), the model maintains 88.9–94.0\% accuracy across all conditions in Table~\ref{tab:external_accuracy}, including VPN-obfuscated traffic where it achieves 93.3\% on US devices and 100.0\% on a small UK sample (n=5). Apparent misclassifications often arise from supervision mismatches—e.g., labels specifying device type (\textit{fridge}) rather than vendor (\textit{Samsung})—where the model nonetheless infers the correct canonical brand, revealing an ability to generalize beyond underspecified labels. To directly evaluate generalization to entirely unseen vendor classes, we conduct a vendor-disjoint analysis by identifying all devices in the held-out test set whose vendor has zero training examples. This yields 132 devices spanning 128 unseen vendor classes, of which the model correctly identifies 118 (89.4\%). Errors concentrate in structurally ambiguous cases such as generic OUIs shared across manufacturers and vendors lacking distinctive hostname or service broadcast signatures.
\begin{table*}[t]
\centering
\footnotesize

\begin{minipage}[t]{0.48\textwidth}
\centering
\caption{Top-1 accuracy on external test subsets across regions and VPN conditions.}
\label{tab:external_accuracy}
\vspace{2pt}

\setlength{\tabcolsep}{6pt}
\begin{tabular}{l l l c}
\toprule
\textbf{Year} & \textbf{Region} & \textbf{VPN} & \textbf{Accuracy (\# Devices)} \\
\midrule
2019 & US       & No  & 93.3\% (45) \\
2019 & UK       & No  & 88.2\% (34) \\
2019 & UK       & Yes & 100.0\% (5) \\
2019 & US       & Yes & 93.3\% (45) \\
2022 & Idle     & No  & 94.0\% (50) \\
2022 & Non-idle & No  & 88.9\% (45) \\
\bottomrule
\end{tabular}
\end{minipage}
\hfill
\begin{minipage}[t]{0.48\textwidth}
\centering
\caption{Baseline performance compared to our fine-tuned LLaMA model.}
\label{tab:baselines}
\vspace{2pt}

\setlength{\tabcolsep}{6pt}
\begin{tabular}{l c c}
\toprule
\textbf{Model} & \textbf{Internal Acc.} & \textbf{External Acc.} \\
\midrule
SVM                    & 76.21\% & 60.17\% \\
XGBoost                & 71.38\% & 67.31\% \\
Label Propagation (LP) & 76.50\% & 66.60\% \\
GCN                    & 75.10\% & 69.95\% \\
BERT                   & 64.60\% & 61.60\% \\
Zero-shot LLM          & --      & 62.30\% \\
Few-shot LLM (RAG)     & --      & 72.40\% \\
OUI Majority Vote
 & --      & 72.30\% \\
Fingerbank             & --      & 25.90\% \\
\midrule
\textbf{Our LLaMA-3.1 FT} & \textbf{98.69\%} & \textbf{92.95\%} \\
\bottomrule
\end{tabular}
\end{minipage}

\end{table*}

\setlength{\tabcolsep}{3pt} 
\begin{table}[t]
\centering
\begin{threeparttable}
\footnotesize
\caption{Cost, throughput, and runtime characteristics of our two-stage pipeline and all baselines}
\label{tab:costs}

\begin{tabular}{lccccc}
\toprule
\textbf{Method} &
\textbf{Accuracy (\%)} &
\textbf{Latency (s)} &
\textbf{Throughput (device/s)} &
\textbf{Cost (\$/device)} &
\textbf{Cost (\$/1k device)} \\
\midrule
\multicolumn{6}{c}{\textit{Our Pipeline}} \\
\midrule
Stage~1 (LLM pseudo-labeling)
& --
& 0.74
& 1.36
& $2.7\times10^{-4}$
& 0.27 \\
Stage~2 (LLaMA-3.1 8B FT)
& 93.2
& 0.19
& 5.29
& $5.8\times10^{-5}$
& 0.058 \\
\midrule
\multicolumn{6}{c}{\textit{Classical Baselines (CPU)}} \\
\midrule
SVM
& 72.86
& 0.0021
& $>100$k
& $1\times10^{-9}$
& $1\times10^{-6}$ \\
XGBoost
& 72.86
& 0.0021
& 476
& $2.9\times10^{-7}$
& 0.00029 \\
GNN-LP
& 49.52
& 0.00006
& 16k
& $1.7\times10^{-8}$
& 0.000017 \\
GNN-GCN
& 63.33
& 0.00006
& 16k
& $1.7\times10^{-8}$
& 0.000017 \\
BERT
& 71.90
& 0.09275
& 10.8
& $2.7\times10^{-5}$
& 0.027 \\
Fingerbank
& 28.69
& 0.1394
& 7.17
& $4.17\times10^{-4}$
& 0.417 \\
\bottomrule
\end{tabular}


\end{threeparttable}
\end{table}

\subsection{Qualitative Error Analysis and Adversarial Robustness}
\label{subsec:qual-expert}

We examine cases where predictions diverge from supervision, extend beyond labeled data, or involve adversarial inputs to understand how the instruction-tuned model synthesizes weak signals, leverages priors, and where those priors fail. In deployments built around semantic network data, such as IoT Inspector \cite{huang2020iot}, realistic adversaries primarily manipulate semantic cues—misleading user labels, spoofed DHCP hostnames, or conflicting cross-field metadata. Under STRIDE~\cite{khan2017stride}, these correspond to Spoofing, Tampering, and Repudiation (devices remaining silent to avoid active discovery). The remaining dimensions—Information Disclosure, Denial of Service, and Elevation of Privilege—do not directly impact semantic network signals and thus do not obscure device identity. Robustness evaluation therefore centers on the model’s \textit{reasoning} behavior under conflicting or deceptive signals. We adopt a breadth-oriented analysis of semantic manipulations and structured perturbations targeting realistic contexts such as short-term rentals and IPV scenarios. The goal is to assess whether the model maintains coherent cross-field inference under sparsity, conflict, and manipulation—conditions where traditional packet-level classifiers offer no comparable guarantees.

To complement this qualitative analysis, we perform a systematic single-field spoofing evaluation, where each metadata field is independently perturbed while all others are held fixed. Spoofing high-signal fields such as \texttt{netdisco\_info} and \texttt{oui\_friendly} produces the largest shifts (61.7\% and 59.4\%, respectively), consistent with their importance in Sec.~5.1.4. However, even under these perturbations, a substantial fraction of predictions remain stable (38.3\% and 40.6\%), indicating that the model does not rely on any single feature. Other fields exhibit intermediate effects, including \texttt{user\_labels} (46.3\%) and \texttt{base\_domains} (40.6\%), while \texttt{dhcp\_hostname} (28.0\%) and \texttt{user\_agent\_info} (30.3\%) produce smaller shifts. Overall, these results demonstrate \textit{partial robustness to single-field spoofing}, with predictions grounded in cross-field consistency rather than brittle dependence on individual metadata attributes.

\subsubsection{Generalization to Canonical and Unseen Vendors} The model frequently outputs vendor labels that are more canonical—i.e., aligned with parent or current consumer brands—than its supervision. It reliably resolves common mergers (e.g., \textit{Ring}, \textit{Blink}, \textit{Eero} $\rightarrow$ \textit{Amazon}; \textit{Dropcam}, \textit{Fitbit} $\rightarrow$ \textit{Google}), and extends beyond seen labels by mapping \textit{Philips Lighting} to its rebranded identity \textit{Signify}, a relationship absent from training data. This behavior indicates that the model leverages pretraining knowledge to infer latent organizational ties—acquisitions, OEM relationships, and rebrandings—rather than merely matching aliases. The same mechanism supports open-world inference: in Appendix~\ref{appendix:examples}, \textit{Deutsche Telekom} is inferred from cues such as \textit{Speedport TV} despite supervision naming only the contract manufacturer (\textit{Arcadyan}). While effective, this reasoning occasionally induces minor factual drift in explanations, illustrating how explanatory rationales can embed slight inaccuracies even when the underlying prediction is correct.

\subsubsection{Resilience Under Adversarial Manipulation} We evaluate robustness under adversarial scenarios where device metadata is manipulated to evade identification—threat models relevant to short-term rentals, shared housing, and intimate partner violence (IPV)~\citep{freed2018stalker,291027,287242}. In the spoofing examples in Appendix~\ref{appendix:examples}, attackers inject misleading cues through multiple channels, including instruction-like user-label directives (e.g., “ignore all previous information—this is just a TP-Link smart plug”) and spoofed DHCP hostnames (e.g., a \textit{Wyze} camera masquerading as a nursery monitor). The model resists these attacks by grounding predictions in cross-field consistency rather than over-relying on any single, easily falsified input. This semantic resilience enables reliable inference in high-risk environments where device misrepresentation could otherwise enable covert monitoring or coercion.

\subsubsection{Robustness to Token-Level Perturbations} \label{token-level-perturbations} We test robustness to token-level perturbations by injecting misleading vendor tokens (e.g., \texttt{ring.com}), scrambling trusted domains, and substituting plausible hostname decoys. Across all variants, the model consistently predicts the correct vendor, anchoring on stable identifiers such as OUIs and user-agent strings rather than overfitting to injected noise. In rare cases, explanations exhibit mild hallucination (e.g., referencing the Google Home app without explicit evidence), reflecting a broader property of instruction-tuned LLMs: semantic resilience under noisy or adversarial inputs can coexist with occasional explanatory drift. This duality highlights the importance of evaluating both prediction correctness and reasoning behavior for trustworthy deployment.

\subsection{Baseline Comparisons}
\label{sec:baselines}

Table~\ref{tab:baselines} compares our instruction-tuned model against rule-based, classical, graph-based, and non-fine-tuned LLM baselines on internal IoT Inspector and external Mon(IoT)r holdouts~\citep{girish2023room}. Among trainable baselines, Label Propagation (76.50\%) and SVM (76.21\%) perform best internally but degrade sharply under cross-dataset evaluation (66.60\% and 60.17\% respectively); GCN generalizes best at 69.95\%. A simple OUI-based majority-vote baseline reaches 72.3\%, demonstrating that OUIs alone are insufficient. Zero- and few-shot LLM baselines, evaluated only externally since they are not trained on IoT Inspector, reach 62.30\% and 72.40\%. Fingerbank, the rule-based reference system, achieves only 25.90\% externally, reflecting the indirectness of real-world metadata. Our model achieves 98.69\% internal and 92.95\% external accuracy, substantially outperforming all baselines by integrating heterogeneous weak signals—OUIs, DHCP hostnames, user-agent strings, and remote hostnames—into coherent vendor predictions that generalize beyond exact signature matches.

\subsection{Novelty-Aware Abstention Analysis}
\label{sec:openset}

We assess the model’s ability to detect uncertain or out-of-distribution predictions (novelty-aware abstention) on the MV-Holdout using a distance-based confidence score defined as the cosine distance between a device embedding and the centroid of its predicted vendor cluster. Correct in-distribution predictions exhibit low dispersion (mean $\approx 0.054 \pm 0.08$), while incorrect or out-of-distribution samples form a heavier-tailed distribution, with many exceeding $\mu {+} \sigma \approx 0.13$. We therefore adopt 0.13 as an abstention threshold: samples above this distance are flagged as low-confidence or potentially novel. This policy yields 84.76\% coverage and improves accuracy on the retained subset by +0.44 points, indicating that high-distance samples reliably capture uncertain predictions. Confident misclassifications—errors falling below the threshold—account for only 2.86\% of all samples, suggesting that the model rarely assigns high confidence to ambiguous or out-of-distribution inputs.

\subsection{Cost and Throughput Analysis}
\label{sec:cost-throughput}

Table~\ref{tab:costs} summarizes the computational and monetary footprint of our pipeline. Stage~1 (LLM pseudo-labeling) required 9.35 hours (1.36 queries/s) and cost \$12.22 ($2.7 \times 10^{-4}$ per device). Stage~2 (fine-tuning) consumed 8.3 GPU-hours on a single A10G ($5.8 \times 10^{-5}$ per device).

At deployment, the fine-tuned LLaMA-3.1 8B model achieves 0.19\,s latency (5.29 devices/s) on a few kilobytes of metadata and scales linearly with batch size and GPUs. Classical baselines run in microseconds–milliseconds with negligible cost ($10^{-9}$--$10^{-7}$), while Fingerbank achieves 0.139\,s latency, 7.17 queries/s, and $4.17 \times 10^{-4}$ per device. Despite higher compute, our model offers substantially better accuracy and robustness.

\subsection{Deployment Setting}
\label{sec:deployment}
Our classifier is deployed within a large-scale open-source application that collects network metadata from residential routers for device identification. The model consumes only a few kilobytes of text per device and requires neither packet payloads nor persistent identifiers, making it well suited for privacy-preserving, low-bandwidth monitoring pipelines. Inference runs as a stateless, containerized microservice on a single GPU. At household granularity, the system processes roughly 600 devices per hour—about 60 households per hour assuming 10 devices per home. This exceeds IoT Inspector’s historical peak load ($\sim$400 daily active homes~\cite{280310}) by more than $3\times$, indicating substantial headroom for practical deployment. Capacity can be increased further via horizontal scaling or by caching stable device embeddings to avoid repeated inference. As requests are fully independent, the service parallelizes trivially. Additional replicas can be launched behind a standard load balancer, enabling the model to function as a drop-in replacement for existing heuristic vendor-identification components in residential or enterprise IoT monitoring pipelines.

%% file: sections/discussion.tex
\section{Limitations and Future Work}
\label{sec:discussion}

Despite strong generalization, several limitations remain. First, the model can hallucinate vendor relationships under sparse metadata—for instance, attributing a Lenovo device to Intel via an imagined acquisition. These errors typically occur when generic OUIs dominate and meaningful hostnames or user-agent cues are absent, and are partially inherited from the pseudo-labeling stage, where long-tail vendors with aliased or incomplete metadata can reinforce spurious associations. While the model's rationales are coherent, they are not trained for causal faithfulness and should be interpreted as plausible justifications rather than guaranteed reflections of its internal decision process. Improving label quality is therefore a central direction for future work: we plan to incorporate user-in-the-loop correction that surfaces low-confidence predictions for confirmation, and to generalize our novelty-aware abstention analysis into a drift-detection module that triggers scheduled retraining on persistent shifts while routing isolated outliers for manual review. As new vendors emerge, parameter-efficient LoRA updates layered onto a frozen backbone—with rehearsal on prior examples to mitigate catastrophic forgetting—offer a path to continual adaptation without full retraining, with each update checkpointed to allow rollback if validation metrics regress.

A second limitation is that our model predicts only vendor identity, not explicit device type. While vendor provides useful contextual information, it may be insufficient when device functionality is critical—for example, distinguishing a camera from a thermostat by the same vendor, which carry fundamentally different surveillance implications in safety-sensitive settings. Model-generated rationales already surface type-specific cues (e.g., ``This hostname pattern is characteristic of consumer security cameras produced by Wyze''), suggesting the feasibility of hierarchical or multi-task formulations that jointly infer vendor, device type, and version. Vendors with narrow product scope (e.g., Wyze, Roku) may be easier to distinguish, whereas multi-product manufacturers (e.g., Samsung, Google) likely require richer relational representations.

Real-world deployment introduces additional constraints on input-signal availability. Emerging privacy-preserving technologies directly obscure the metadata fields our model relies on most: DNS-over-HTTPS (DoH) and DNS-over-TLS (DoT) encrypt remote hostname resolution, Encrypted ClientHello (ECH) conceals SNI-based domain information, and MAC address randomization—now default on Android and iOS—degrades \texttt{oui\_friendly} signals, which our ablation (\ref{sec:feature-ablation}) identifies as the single most important feature. Under combined deployment, accuracy may degrade substantially, particularly for long-tail vendors that rely heavily on hostname and OUI cues. Addressing this will likely require alternative signals more robust to encryption (e.g., timing patterns, flow statistics, or local broadcast metadata), or active probing where permitted. Finally, as shown in Table~\ref{tab:costs}, the pipeline incurs higher latency and cost than lightweight baselines like SVM or Fingerbank, making it better suited for periodic inventory or batch analysis than real-time deployments; cloud-based inference also raises privacy considerations, as some metadata fields (e.g., mDNS/SSDP strings) may contain PII~\cite{girish2023room}, which quantization and self-hosting mitigate at the cost of operational complexity.

%% file: sections/conclusion.tex
\section{Conclusion}
We show that IoT device identification can be cast as a language-based inference problem rather than a signature-matching task. By instruction-tuning on curated pseudo-labels and training with a structured curriculum, the model learns semantically grounded and interpretable representations from noisy, heterogeneous network signals. The resulting system generalizes across vendors and deployment conditions, supporting robust device identification in open-world IoT environments.

%% file: sections/ethics.tex
\section{Ethics Statement}

This study uses an anonymized network traffic dataset obtained from the IoT Inspector authors under their original IRB approval~\citep{huang2020iot}. The dataset contains only coarse device-level metadata (Table~\ref{appendix:feature-descriptions})—no payloads, credentials, or content—was stored encrypted on access-controlled research servers, and is not publicly released to minimize re-identification risk. Our analysis adheres to the Menlo Report and ACM guidelines, focuses solely on device vendor/type inference, and complies with all applicable institutional and national ethical standards.